# A unified three-branch deep learning framework for automated detection and counting of wildlife in aerial imagery

Zeyu Xu[1], Tiejun Wang[1], Andrew Skidmore[2], Megan Claase[3], Harry Wells[4,5], Isla Duporge[4], Ramiro D. Crego[6], Richard Lamprey[2]

[1] School of Geospatial Engineering and Science, Sun Yat-Sen University, Zhuhai, China

[2] Faculty of Geo-Information Science and Earth Observation, University of Twente, Enschede, The Netherlands

[3] Boma and Badingilo National Parks, African Parks, Juba, South Sudan

[4] Department of Ecology and Evolutionary Biology, Princeton University, Princeton, USA

[5] Lolldaiga Hills Research Programme, Nanyuki, Kenya

[6] School of Biological, Earth & Environmental Sciences, Environmental Research Institute, University College Cork, Cork, Ireland

**Abstract**

Automating aerial wildlife surveys requires reliable animal detection, consistent delineation of observation strips, and correction for repeated coverage between successive images. We present a unified three-branch deep learning framework that integrates these tasks through WildSeg, WildMatch and WildDet. WildSeg combines rod segmentation with geometric refinement to delineate counting strips from aircraft-mounted markers. WildMatch uses teacher–student distillation to learn dense image correspondences and identify overlapping survey areas without relying on geolocation metadata. WildDet combines dual-scale detection with feature fusion, density supervision and empty-image gating to support animal detection and counting. We evaluated the framework using oblique aerial imagery of four antelope species from the Great Nile Migration Landscape in South Sudan. WildSeg achieved reliable rod extraction, while WildMatch approached teacher-level accuracy with a substantially lighter and faster model. Both modules maintained high accuracy on resized images, improving processing efficiency. WildDet improved detection performance across all four species and reduced detection-based count MAE, RMSE and mean bias by 21.0%, 4.6% and 27.8%, respectively, relative to YOLO11. By connecting spatial survey constraints with animal detection, the framework provides a practical and computationally efficient approach to processing large aerial image collections for wildlife monitoring and population surveys.

## 1. Introduction

Deep learning has shown considerable promise in advancing wildlife monitoring through manned aircraft, satellite, and drone imagery (Delplanque et al. 2023a; Eikelboom et al. 2019;

Kellenberger et al. 2018; Kumar et al. 2024; Tuia et al. 2022; Wu et al. 2023b). However, since the primary goal of aerial wildlife surveys is not image interpretation (e.g., object detection) but accurate estimation of animal population sizes, a gap remains between detection models and abundance estimates (African Parks 2023; Frederick et al. 2010).

Systematic Reconnaissance Flight (SRF) is one of the most widely used methods for aerial wildlife and livestock surveys across Africa (African Parks 2023; Delplanque et al. 2023b; Frederick et al. 2010; Lamprey et al. 2020a; Norton-Griffiths 1978; Ogutu et al. 2016; PAEAS 2014). SRF combines systematic transect flights with population estimation models (Jolly 1969) to estimate species abundance across large landscapes. Recent advances, including the integration of high-definition oblique camera systems, have improved the accuracy and consistency of SRF-based surveys, enabling more reliable multi-species population estimates (Lamprey et al. 2020b). In aerial imagery, not all detected animals are included in the final count—targets outside the counting strip or within overlapping areas between images must be excluded to avoid overestimation (Delplanque et al. 2024). However, to the best of our knowledge, no existing framework has automatically addressed these issues.

Traditional aerial wildlife surveys in Africa often include visual markers, typically rigid rods mounted on the aircraft's wing struts, to define the counting strip within which observations are recorded. This method is commonly used by human 'rear-seat observers' (RSOs), who visually spot herds within the sample strips, estimate their numbers, and then photograph them for later validation (Bussière and Potgieter 2023; Craigie et al. 2010; Lamprey et al. 2020a, 2020b; Frederick et al. 2010). In current aerial survey practices, the rear-seat observer (RSO) method and the oblique-camera-count (OCC) system (Lamprey et al. 2020a, 2020b) are often employed simultaneously, typically acquiring tens or even hundreds of thousands of images with visible counting strips (African Parks 2023). In such survey designs, fixed-mounted OCC cameras are carefully aligned with the RSOs' field of view, allowing for direct comparison and validation between photographic and visual observations (African Parks 2023; Lamprey et al. 2020b). The extraction of counting strips in large image datasets is still not integrated into automated pipelines and remains mostly manual (Delplanque et al. 2024). Due to variations in aircraft attitude, the rods in the images are typically not parallel to the image boundaries and are often affected by sunlight reflections, which introduce additional challenges. To address this, we explore the use of deep learning-based semantic segmentation networks (Badrinarayanan et al. 2017; Long et al. 2015; Ronneberger et al. 2015) to detect the rods and extract the counting strip based on geometric relationships. This study presents the first automated implementation of this process.

In camera-count systems, cameras are often set to automatically capture images at regular intervals (e.g., every 2 seconds; African Parks 2023), with the aim to generate overlapping regions between successive aerial images. However, variations in aircraft speed (e.g., due to wind conditions) can lead to inconsistent overlap between images—slower speeds increase overlap, while faster speeds reduce it. In extreme cases, high speeds may result in no overlap at all, leaving unsurveyed gaps along the transect (African Parks 2023). This variability further complicates efforts to correct for duplicate detections. If not addressed, animals appearing in these areas may be double counted, leading to significant biases in population estimates. While some datasets, such as the Aerial Elephant Dataset (Naude and Joubert 2019), include overlapping frames, they do not explicitly differentiate between original and duplicate annotations. A few studies have acknowledged the problem of duplicate detections, but most rely on manual correction rather than automated solutions (Delplanque et al. 2024).

Geospatial coordinates associated with aerial imagery can be used to delineate overlaps; however, in resource-constrained settings, such coordinates recorded during aerial surveys with manned aircraft are often insufficiently accurate or even entirely unavailable for this task (Delplanque et al. 2024), necessitating image matching approaches based solely on optical features. Identifying overlap automatically in the absence of geospatial information is particularly challenging in natural settings, due to the visual redundancy of the landscape. The variability in overlap caused by fluctuating aircraft speeds further exacerbates this challenge, as it prevents the assumption of a consistent overlap percentage between images. Unlike urban or agricultural scenes, where distinct features such as buildings or field boundaries support precise matching, savanna environments are more challenging, being characterized by repetitive and less distinct patterns formed by grasses, shrubs, and scattered trees.

Classical keypoint-based image matching algorithms such as Scale-Invariant Feature Transform (SIFT; Lowe 2004) and Oriented FAST and Rotated BRIEF (ORB; Rublee et al. 2011) are computationally efficient but often struggle with variations in scale, illumination, and viewpoint. Deep learning-based approaches, including SuperPoint (DeTone et al. 2018), D2-Net (a trainable CNN for joint detection and description of local features; Dusmanu et al. 2019), and Local Feature Transformer (LoFTR; Sun et al. 2021), have shown improved robustness by learning local feature descriptors in an end-to-end fashion. Nevertheless, their robustness remains limited in scenes dominated by low-contrast, repetitive textures such as those commonly encountered in aerial wildlife imagery. This underscores the need for new strategies that leverage landscape-specific visual features to enhance overlap detection in wildlife surveys.

Beyond these technical issues, the ecological setting presents its own complexities. Large grasslands and savannas across the world support high densities of migratory herbivores, posing unique challenges for automated monitoring (Cuni-Sanchez et al. 2016; Purdon et al. 2018). Wildlife inhabiting these ecosystems frequently exhibit colorations and body shapes that blend into the surrounding environment, complicating detection (Mondal 2020). Illumination variability, occlusion, and object scale further reduce the effectiveness of standard detection architectures, especially in the absence of ground truth references. These factors must be accounted for when applying neural networks for wildlife detection. The YOLO (You Only Look Once; Redmon et al. 2016) family of object detectors has been extensively applied to wildlife detection in aerial imagery (Devi Priya et al. 2022; Petso et al. 2021; Wu et al. 2023a; Xie et al. 2023; Xu et al. 2024). Prior studies have demonstrated that YOLO models can achieve performance comparable to fully supervised methods, even when trained on coarse annotations such as point labels or loosely defined bounding boxes (Xu et al. 2025). However, current YOLO-based methods have not fully considered the contrast between animal targets and their backgrounds. In addition, while YOLO adopt multi-scale detection structures, their design is not well aligned with the actual scales of animals in aerial imagery (Jocher et al. 2024; Redmon and Farhadi 2018).

Despite the relevance of these challenges, no existing study offers integrated, automated solutions for estimating wildlife populations from aerial photography. In this study, we introduce a unified, three-branch deep learning framework specifically designed to meet the operational and environmental constraints of aerial wildlife surveys.

We introduce WildSeg, WildMatch and WildDet to address these tasks. WildSeg combines lightweight convolutional blocks and learnable offset sampling with geometric rod-boundary fitting. WildMatch learns dense image correspondences from a RoMaV2 teacher. WildDet combines detection at two feature scales with density estimation and an empty-image gate. The networks are trained separately; the framework is a staged workflow rather than a jointly optimized three-network model. We evaluate rod detection, overlap extraction and animal detection at the component level.

## 2. Study area and dataset

This study utilizes data collected during the 2023 systematic reconnaissance flight (SRF) aerial survey targeting antelope populations in the Great Nile Migration Landscape, South Sudan (African Parks, 2023). The survey involved systematic flights along parallel transects and provided population estimates across large spatial extents. Similar sampling designs have been operational in Africa for decades, including Kenya (Lamprey et al., 2020b) and Uganda (Lamprey et al., 2020a).

The survey was conducted using two fixed-wing, four-seater Cessna 182 aircraft, each equipped with high-resolution Sony α7 IV and α7R IV cameras (African Parks 2023). The cameras were positioned to replicate the RSOs' fields of view and were set to automatically capture photographs along the flight line at 2-second intervals. Approximately 330,000 images were captured in total, with around 250,000 taken on-transect (African Parks 2023). As illustrated in Figure 1, two rigid rods were mounted on the wing struts of each aircraft to define the counting strip visible in each image. In accordance with standard aerial survey protocols (African Parks 2023), only animals located between the rods were included in the official counts. The extent of overlap between successive images was influenced by the aircraft's speed and altitude above ground level.

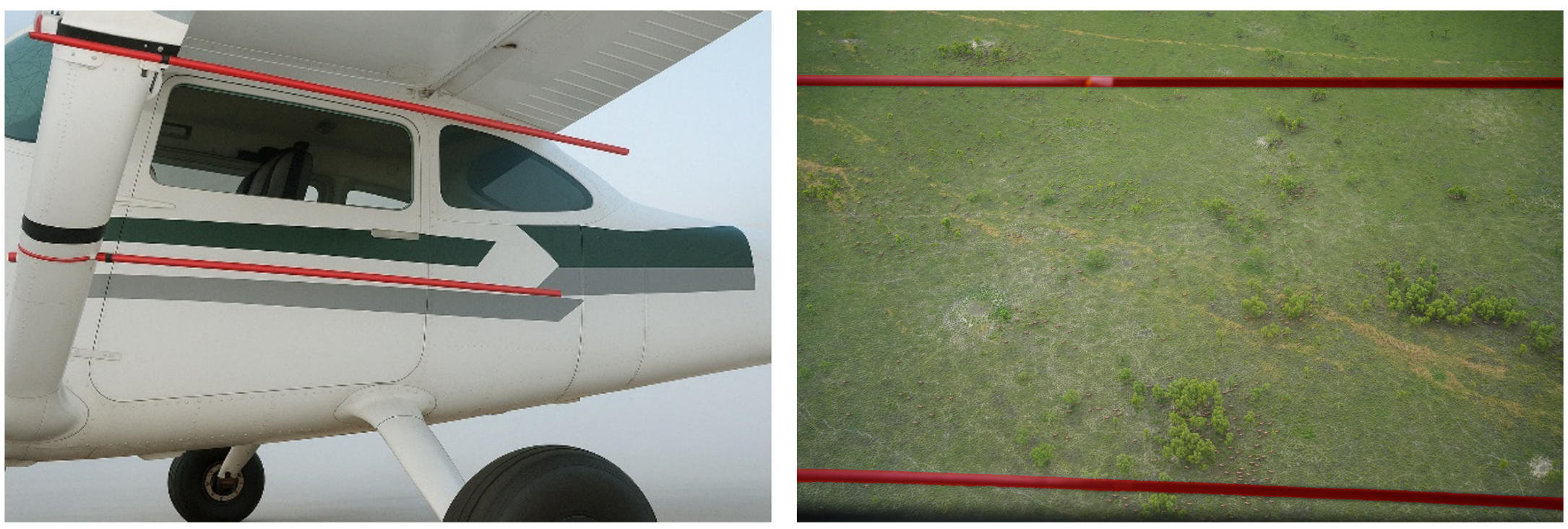

**Figure 1.** Configuration of counting rods and aerial observation strip. The left panel shows parallel rigid rods (highlighted in red) mounted on the wing strut of a Cessna aircraft, which define the lateral boundaries of the counting strip during aerial surveys. The right panel presents an aerial view through the camera lens, showing the ground area visible between the upper and lower rods. Only animals located within this defined strip are included in official counts, in accordance with standard survey protocols (African Parks, 2023; Frederick et al., 2010; Norton-Griffiths, 1978).

For this study, we used a subset of typical images (n = 2,065) with corresponding human annotations, including 1,778 images at 7008×4672 pixels and 287 images at 9504×6336 pixels. The spatial extent of our study area is shown in Figure 2. Our analysis focuses on four key antelope species that dominate the study area (African Parks 2023): white-eared kob (Kobus kob leucotis), tiang (Damaliscus lunatus tiang), Bohor reedbuck (Redunca redunca), and Mongalla gazelle (Eudorcas albonotata). The white-eared kob is a highly migratory species endemic to Central Africa's floodplains and savannas (Fryxell and Sinclair 1988; Marjan 2014). The tiang is a robust, swift-moving species that prefers open grasslands (Cotterill 2003; Marjan 2014). The Bohor reedbuck typically inhabits tall grass near water bodies and exhibits characteristic bounding movements in response to disturbance (Djagoun et al. 2013). The Mongalla gazelle is a smaller antelope distinguished by a prominent dark lateral stripe, which helps differentiate it from other species yet enhances its cryptic concealment within the environment (Kingdon and Hoffmann 2013; Schlitter et al. 2014). In addition to the four focal species, an auxiliary "Others" category includes a small number of instances of other animal species and people visible in the imagery. This category was included to help the model distinguish the focal species from other targets and was not treated as a focal category in the species-specific analyses.

The dataset was divided into a training set of 1,564 full-resolution images and a test set of 501 full-resolution images. Temporal continuity was preserved within each subset to support overlap analysis between adjacent frames. The distribution of annotated targets across the five categories in the training and test sets is presented in Table 1.

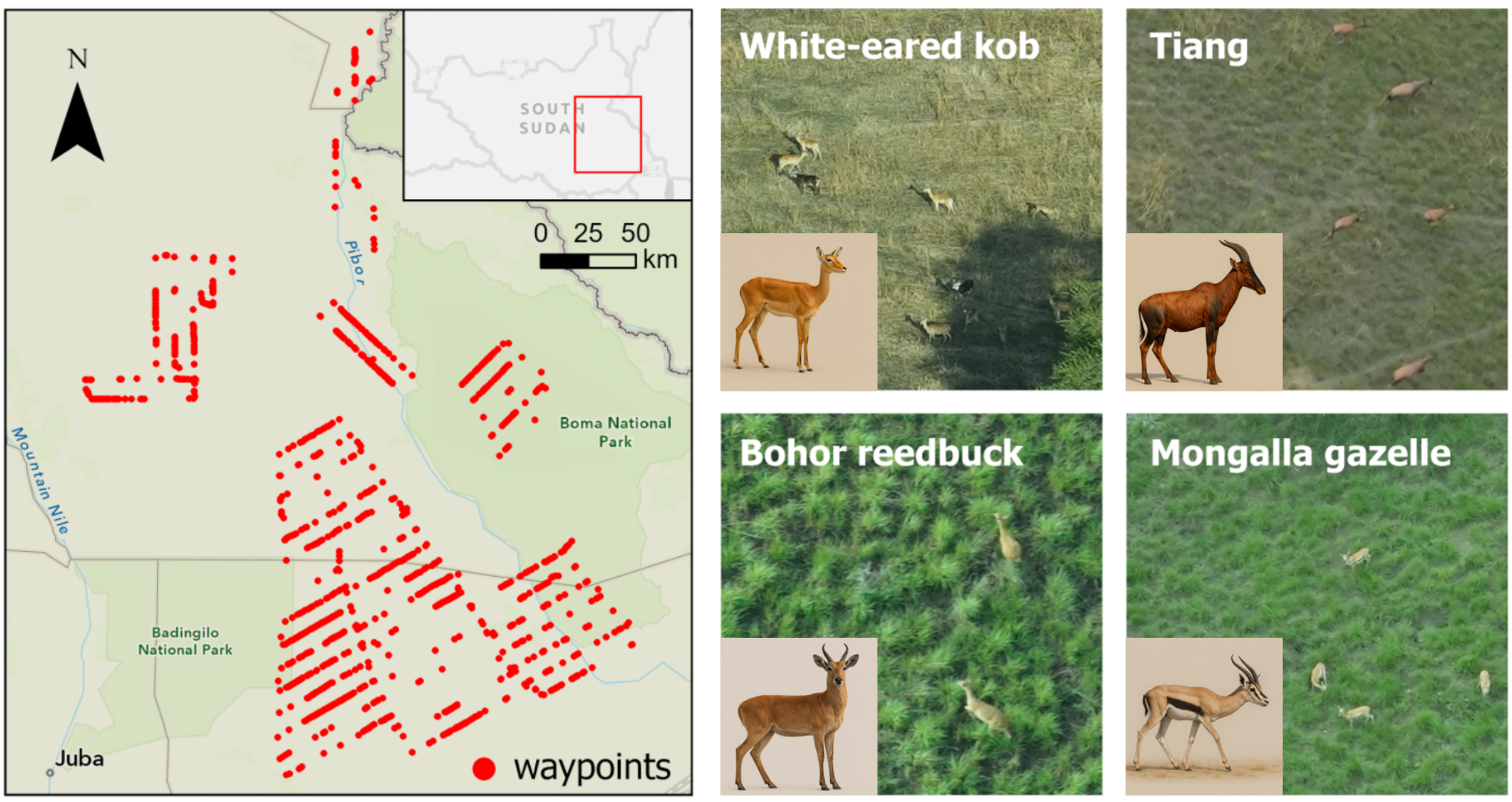

( a ) ( b )

**Figure 2.** Overview of the study area and representative imagery of target species. (a) Geographic extent of the study area in eastern South Sudan, encompassing key regions within and adjacent to Boma and Badingilo National Parks. Red dots indicate waypoints along transects (the subset we used), representing general flight paths rather than exact image acquisition locations. These points were generated by the aircraft's automatic recording system to document the flight trajectory. The inset map provides the national context of the survey area within South Sudan. (b) Representative aerial imagery of the four antelope species surveyed: white-eared kob, tiang, Bohor reedbuck, and Mongalla gazelle. These examples highlight variation in species morphology, group formation, and background texture, all of which contribute to the complexity of automated detection in savanna environments.

Table 1. Numbers of annotated instances by category in the training and test sets.

| **Species** | **Training set** | **Test set** |
|---|---|---|
| White-eared kob | 28,426 | 6,922 |
| Tiang | 5,022 | 1,393 |
| Bohor reedbuck | 2,673 | 853 |
| Mongalla gazelle | 1,814 | 409 |
| Others | 385 | 605 |

## 3. Method

### 3.1. Overall framework structure

The proposed framework comprises three complementary modules, WildSeg, WildMatch and WildDet, that connect spatial survey constraints with animal detection and counting (Figure 3). WildSeg and WildMatch process resized images for efficient strip delineation and overlap extraction, while image patches provide the inputs for animal detection. The three modules are trained separately, and their outputs are combined within a common survey workflow.

WildSeg extracts aircraft-mounted rods and applies geometric refinement to generate a counting-strip inclusion mask for each image. WildMatch establishes dense correspondences between adjacent images and estimates a homography to project the preceding strip into the current frame. The intersection of the projected strip and the current strip defines the overlap to be excluded. WildMatch learns from a RoMaV2 teacher through knowledge distillation during training; inference uses the student model.

WildDet predicts animal bounding boxes and species labels, together with density-based count estimates and an animal-presence probability. For detection-based survey counting, predicted boxes are mapped to the original image coordinates and filtered using the spatial masks. Only detections whose centers fall within the current counting strip and outside the overlap region are retained. This integration produces animal counts constrained by both the observation boundaries and repeated image coverage.

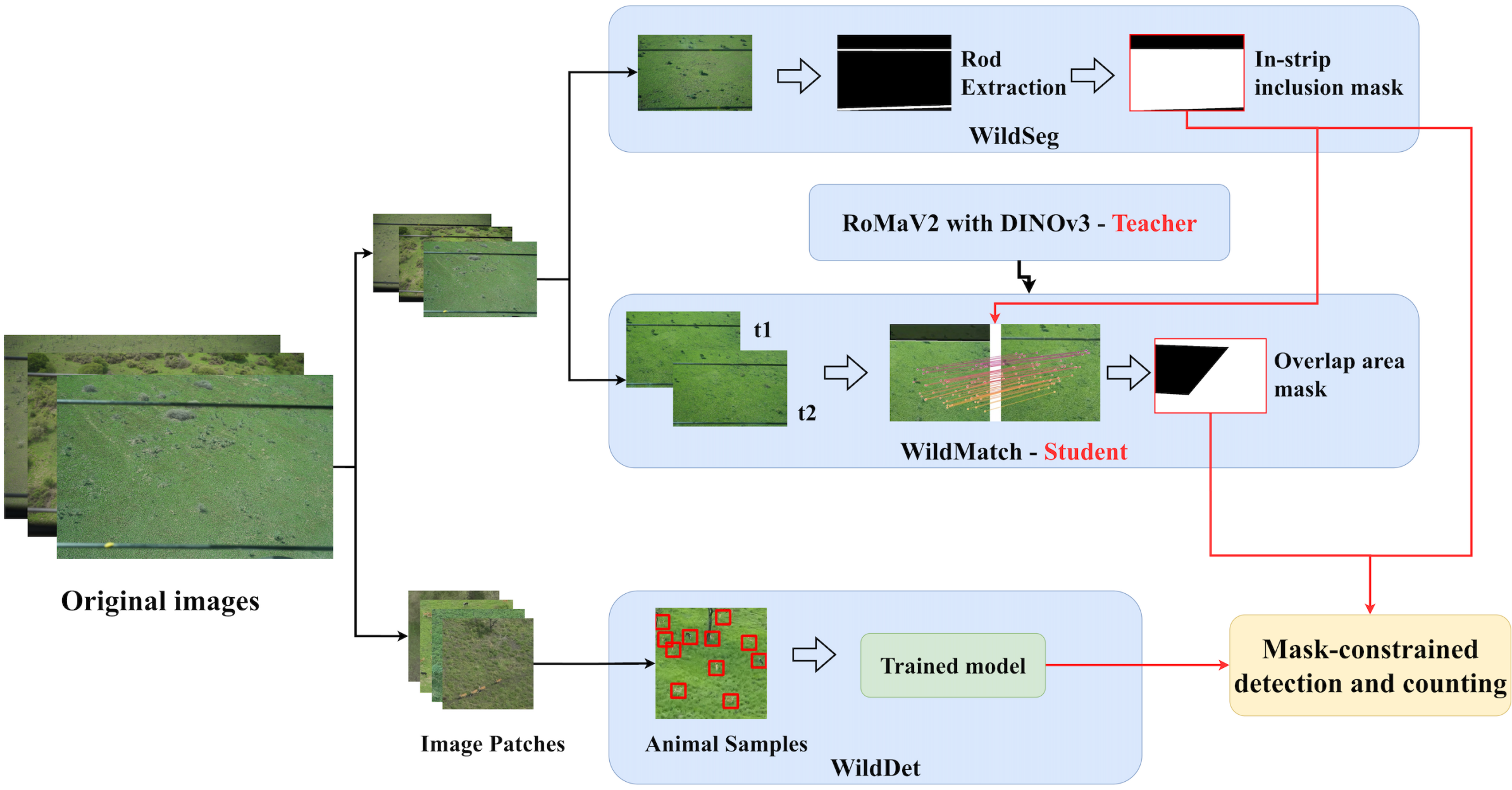

**Figure 3.** Overall structure of the proposed framework for aerial wildlife detection and counting. WildSeg extracts rods from resized images to generate counting-strip inclusion masks. WildMatch, trained through distillation from a RoMaV2 teacher, estimates correspondences between adjacent images to identify overlapping counting areas. WildDet detects and classifies animals in image patches. The strip and overlap masks are combined with the detection results to retain animals within the valid counting region. Teacher supervision is used only during training.

### 3.2. WildSeg for counting strip extraction

WildSeg uses a U-Net-style encoder–decoder with four downsampling stages, four upsampling stages and same-scale skip connections. Encoder channels increase from 64 to 128, 256 and 512, followed by a 1024-channel bottleneck. Decoder channels decrease through 512, 256, 128 and 64. Depthwise separable convolutions combine a $3 \times 3$ depthwise operation with a $1 \times 1$ pointwise operation, batch normalization and ReLU. A final $1 \times 1$ convolution and sigmoid produce rod probabilities.

Learnable offset-sampling modules in the encoder accommodate elongated rod structures. Horizontal and vertical offsets are bounded using tanh and used for bilinear feature sampling. The sampled features are averaged, combined with a conventional $3 \times 3$ convolutional branch, and added residually to the encoder features. The first two stages use seven sampling groups and the next two use five. The inference configuration enables offset sampling and does not enable attention gates.

Images are first resized while preserving aspect ratio to a maximum long edge of 1024 pixels and then resized to $512 \times 512$ for segmentation. Rod probabilities are thresholded at 0.75. Connected components spanning at least 80% of the image width are retained, and upper and lower rod candidates are selected according to their vertical positions and horizontal spans. A quadratic curve is fitted to the lower boundary of the upper rod and the upper boundary of the lower rod. The region between the curves is filled and mapped back to the original image size using nearest-neighbor interpolation.

When only the upper rod is found, the area below it is retained; when only the lower rod is found, the area above it is retained. Failure to identify either rod marks the frame as unavailable for strip-based counting. Constant or interpolated boundaries provide fallbacks when curve fitting lacks sufficient samples or fails.

### 3.3. WildMatch for overlap extraction

WildMatch estimates dense correspondences directly from an image pair. A shared ResNet18 descriptor truncated at layer3 provides stride-16, 256-channel features. Global cosine similarities are scaled by a learnable temperature and normalized using softmax. Probability-weighted target coordinates form the initial correspondence field. A convolutional head predicts coordinate residuals and confidence logits.

Three refinement stages use a separate VGG19-BN feature pathway at strides 4, 2 and 1. Projected feature dimensions are 192, 48 and 12, respectively. The first two refinement stages use local correlation neighborhoods of 7 × 7 and 3 × 3; the final stage uses no local correlation. Each stage combines source features, sampled target features and displacement information to update coordinates and confidence.

The student is trained by distillation from the precise configuration of RoMaV2 (Edstedt et al. 2025). Teacher correspondence fields and soft confidence probabilities supervise the coarse matcher and all three refinement stages. The objective combines confidence-weighted coordinate error, confidence binary cross-entropy and a corner projection term:

$$L_{match} = \sum_{s} \left(10L_{warp,s} + L_{conf,s}\right) + 0.01L_{corner} \quad (1)$$

Coordinate error sums absolute differences over the two coordinate dimensions and normalizes by total teacher confidence. Corner supervision compares four corner correspondences with the teacher homography when available. The ResNet18 and VGG19-BN feature extractors use ImageNet initialization during training. Image-pair partitions group acquisition sessions, keeping corresponding left- and right-camera directories within the same subset.

### 3.4. Animal object detection

WildDet uses convolutional downsampling, C3k2/C3k blocks, SPPF and C2PSA in its backbone. Coordinate attention is applied to P2, P3 and P4 features. The neck propagates information from P5 to P3 and then back to P4, combining backbone and neck features with non-negative, normalized learnable weights. Detection heads operate at P3 and P4 with strides of 8 and 16. P5 remains in the backbone for contextual features and presence gating, but has no detection output.

Bounding-box regression represents each edge using 16 discrete bins. Classification uses depthwise and pointwise convolutions. A local classification option supplies backbone P3/P4 features to the classification heads while regression uses fused features. This separates the classification feature path from top-down fusion.

A density branch projects P3 and upsampled P4 features to a common channel dimension, concatenates and fuses them, and predicts non-negative density maps at stride 8. In species-specific mode, each class has one density channel. Summing a channel gives its continuous count estimate. Training targets place a Gaussian at each annotated box center. Its standard deviation is max(0.8, b/(4s)), where b is the box long-edge length and s is the density stride. Each Gaussian is normalized within the image to integrate to one.

An empty-image gate applies global average pooling to P5, followed by layer normalization and a two-layer multilayer perceptron. Its target indicates whether the augmented input contains any valid animal annotation. When enabled at inference, a probability below the chosen threshold skips the neck, detection head and density head, returning zero detections and density. The backbone still runs. For sliding-window inputs, the gate operates on individual crops.

The training objective combines detection, count, density and gating losses:

$$L_{det} = 7.5L_{CIoU} + 0.5L_{cls} + 1.5L_{DFL} + L_{count} + 10L_{density} + 0.5L_{gate} \tag{2}$$

Classification uses binary cross-entropy with soft targets, and regression uses complete IoU and distribution focal losses. The count term is Smooth L1 between density integrals and annotation counts. Density supervision uses the summed squared error per image. Gate supervision balances positive and empty groups within a batch when both are present. Detection target assignment uses classification and IoU alignment, with up to ten positive assignments per ground-truth object.

### 3.5. Accuracy assessment

#### 3.5.1 Evaluation of counting strip extraction

Rod detection accuracy is the number of correctly detected reference rods divided by the total number of reference rods. A rod is considered correct when the predicted rod pixels are uninterrupted and its width remains within ±5% of the reference width. This object-level criterion emphasizes structural continuity and width consistency, rather than pixel-wise overlap alone. WildSeg and U-Net are compared using this criterion.

#### 3.5.2 Evaluation of overlap area extraction

Overlap extraction is assessed through acquisition failures and incorrect extracted regions. An acquisition failure occurs when an image pair expected to overlap does not yield an accepted region because it fails the criterion of at least 50 inliers and a mean direction angle no greater than 20°. A region error occurs when a region is produced but is judged incorrect. Both error rates use the total number of evaluated image pairs as their denominator:

$$E_{total} = \frac{N_{failure}+N_{incorrect}}{N_{pairs}}, \;\; Accuracy = 1 - E_{total}, \tag{3}$$

The student and teacher are evaluated with the same definitions.

3.5.3 Animal detection and counting evaluation

Animal detection and counting performance were evaluated using complementary metrics. Detection performance was assessed using average precision (AP) and the best F1-score under both class-aware and class-agnostic evaluation. Class-aware evaluation considers both object localization and species assignment, whereas class-agnostic evaluation treats all animal categories as a single class. Species-level AP was also reported for Kob, Gazelle, Tiang and Reedbuck.

AP summarizes the precision–recall curve across detection confidence thresholds. Precision measures the proportion of predicted detections that are correct, while recall measures the proportion of reference animals successfully detected. The F1-score is the harmonic mean of precision and recall:

$$\mathrm{F1} = \frac{2 \cdot \mathrm{Precision} \cdot \mathrm{Recall}}{\mathrm{Precision} + \mathrm{Recall}}$$

$$\text{where Precision} = \frac{TP}{TP+FP}, \quad \text{Recall} = \frac{TP}{TP+FN} \tag{4}$$

where TP, FP, and FN represent true positives, false positives, and false negatives, respectively.

Counting performance was assessed using mean absolute error (MAE), root mean squared error (RMSE) and mean bias. Predicted counts were obtained from the number of detected bounding boxes, rather than from density-map integrals. Let $p_i$ and $y_i$ denote the predicted and reference animal counts for image $i$, respectively, and let $N$ be the total number of evaluated images. MAE measures the average absolute count discrepancy:

$$\mathrm{MAE} = \frac{1}{N}\sum_{i=1}^{N}|p_i - y_i| \tag{5}$$

RMSE gives greater weight to larger count errors:

$$\mathrm{RMSE} = \sqrt{\frac{1}{N}\sum_{i=1}^{N}(\mathrm{p}_i - \mathrm{y}_i)^2} \tag{6}$$

Mean bias measures the average signed count error, with positive values indicating overcounting and negative values indicating undercounting:

$$\text{Mean bias} = \frac{1}{N}\sum_{i=1}^{N}(\mathrm{p}_i - \mathrm{y}_i) \tag{7}$$

Species-level RMSE was calculated using the predicted and reference counts for each corresponding species. Together, these metrics characterize detection quality, counting error and the direction of counting bias.

## 4. Experimental settings

To evaluate the effectiveness of the proposed three-branch framework, we conducted experiments on each component using prepared datasets and consistent hardware resources. All experiments were executed on a Linux-based system (Ubuntu 20.04.6 LTS) equipped with an NVIDIA A100 GPU (80 GB, PCIe), ensuring sufficient memory and computational capacity to handle large-scale image processing and deep learning workflows. The experimental settings for the three branches are as follows, with all parameters selected as the optimal values after multiple preliminary tests.

### 4.1 Counting strip extraction

Counting strip extraction used the image preprocessing procedure described in Section 3.2. A subset of 100 images was randomly selected from the training set for model training, while all 501 test images were used for evaluation.

### 4.2 Overlap area extraction

WildMatch was trained using pairs of aerial images, with correspondence fields and confidence probabilities provided by a RoMaV2 teacher model. Image pairs were partitioned by acquisition session, with corresponding left- and right-camera directories assigned to the same subset.

### 4.3. Animal object detection

The training set included 1,564 full-resolution images, from which 11,658 patches of 512 × 512 pixels were extracted. The test set comprised 3,047 patches of the same size extracted from the 501 full-resolution test images. All WildDet detection and counting metrics were evaluated on these test patches.

## 5. Results

### 5.1 In-strip area extraction results

WildSeg achieved 99.7% rod detection accuracy, compared with 93.8% for U-Net (Table 2), representing a 6.3% improvement. These values describe the proportion of rods meeting the continuity and width criteria, rather than pixel-wise F1 or the accuracy of the final counting-strip masks. Figure 4 compares rod segmentation results from U-Net and WildSeg in two aerial images with different background conditions. U-Net produces fragmented predictions along the upper rod and misses part of the lower rod in Figure 4 (a), while a gap remains in its upper-rod prediction in Figure 4 (b). WildSeg recovers both rods more continuously across the image width in both examples. These results illustrate its ability to preserve elongated rod structures against varying vegetation backgrounds, providing coherent boundaries for subsequent counting-strip delineation.

Table 2. Rod detection accuracy under the continuity and width criteria.

| Model | Rod detection accuracy |
| --- | --- |
| U-Net | 93.8% |
| WildSeg | 99.7% |

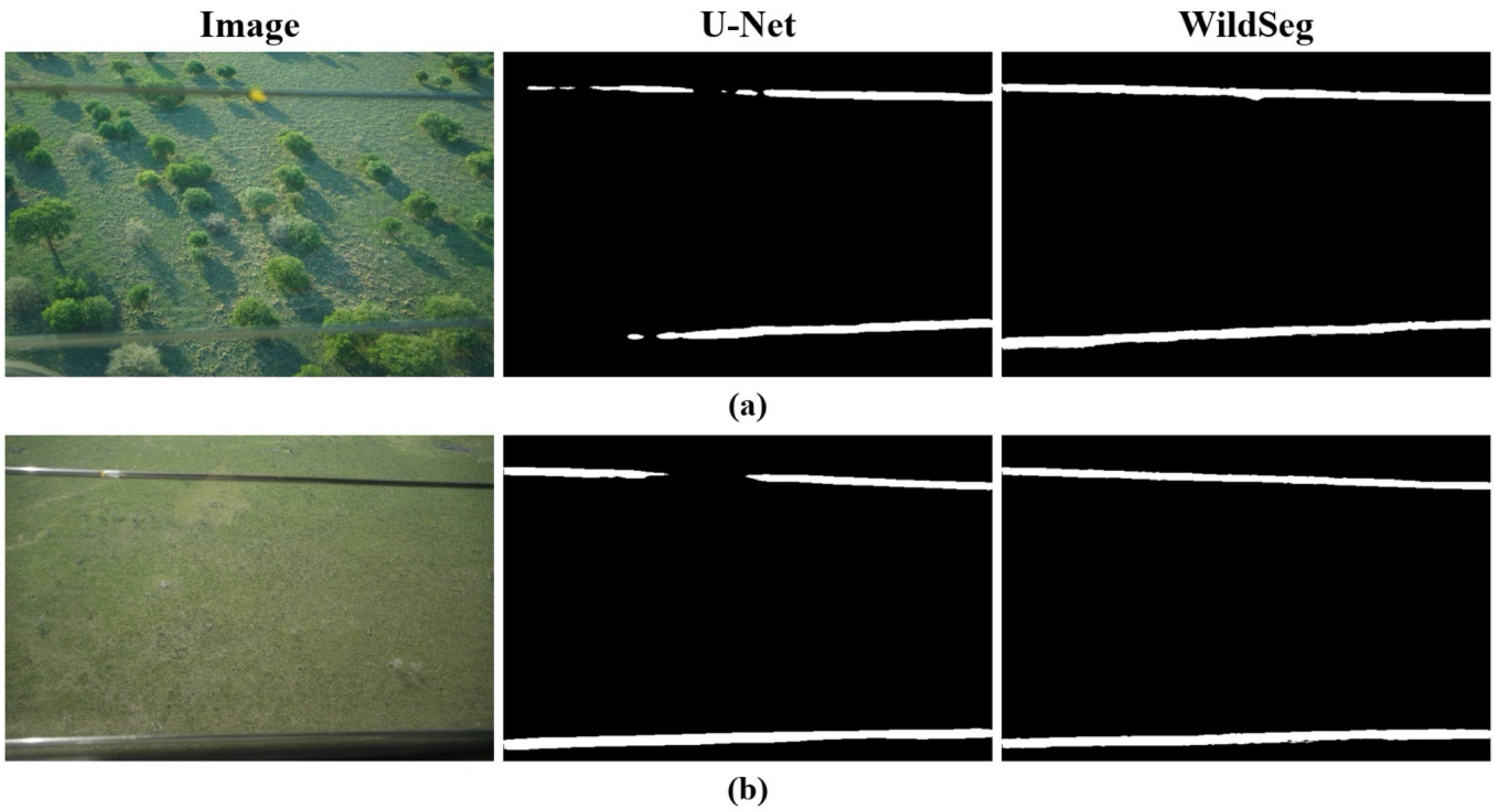

**Figure 4.** Qualitative comparison of rod segmentation using U-Net and WildSeg. Rows (a) and (b) show examples with heterogeneous woody vegetation and a more uniform grassland background, respectively. Columns show the original image, U-Net prediction and WildSeg prediction. White pixels indicate predicted rod regions, and black pixels indicate background. WildSeg produces more continuous rod predictions with fewer gaps and missing segments.

### 5.2 Non-overlap area extraction results

WildMatch had an acquisition failure rate of 1.96% and a region error rate of 0.10%, giving a total error rate of 2.06% and an accuracy of 97.94% (Table 3). The teacher had corresponding error rates of 1.65%, 0.05% and 1.70%. Applying Equation (3) to the teacher total error gives an accuracy of 98.30%. Acquisition failures accounted for most errors in both models. Figure 5 illustrates the overlap extraction procedure between two successive aerial images. WildMatch establishes correspondences within the counting strips, and the retained matches are used to estimate a homography. This transformation projects the footprint of the preceding image into the current image, identifying their shared coverage. For survey counting, the projected preceding strip is intersected with the current strip to define the overlap region to be excluded. This procedure translates image correspondences into an explicit spatial mask for preventing repeated counting across successive frames.

Table 3. Overlap extraction performance of WildMatch and its teacher under the same evaluation criteria. Accuracy is calculated as 100% minus total error rate.

| **Metric** | **WildMatch** | **Teacher** (RoMaV2 precise) |
|---|---|---|
| Acquisition failure rate | 1.96% | 1.65% |
| Region error rate | 0.10% | 0.05% |
| Total error rate | 2.06% | 1.70% |
| Accuracy | 97.94% | 98.30% |

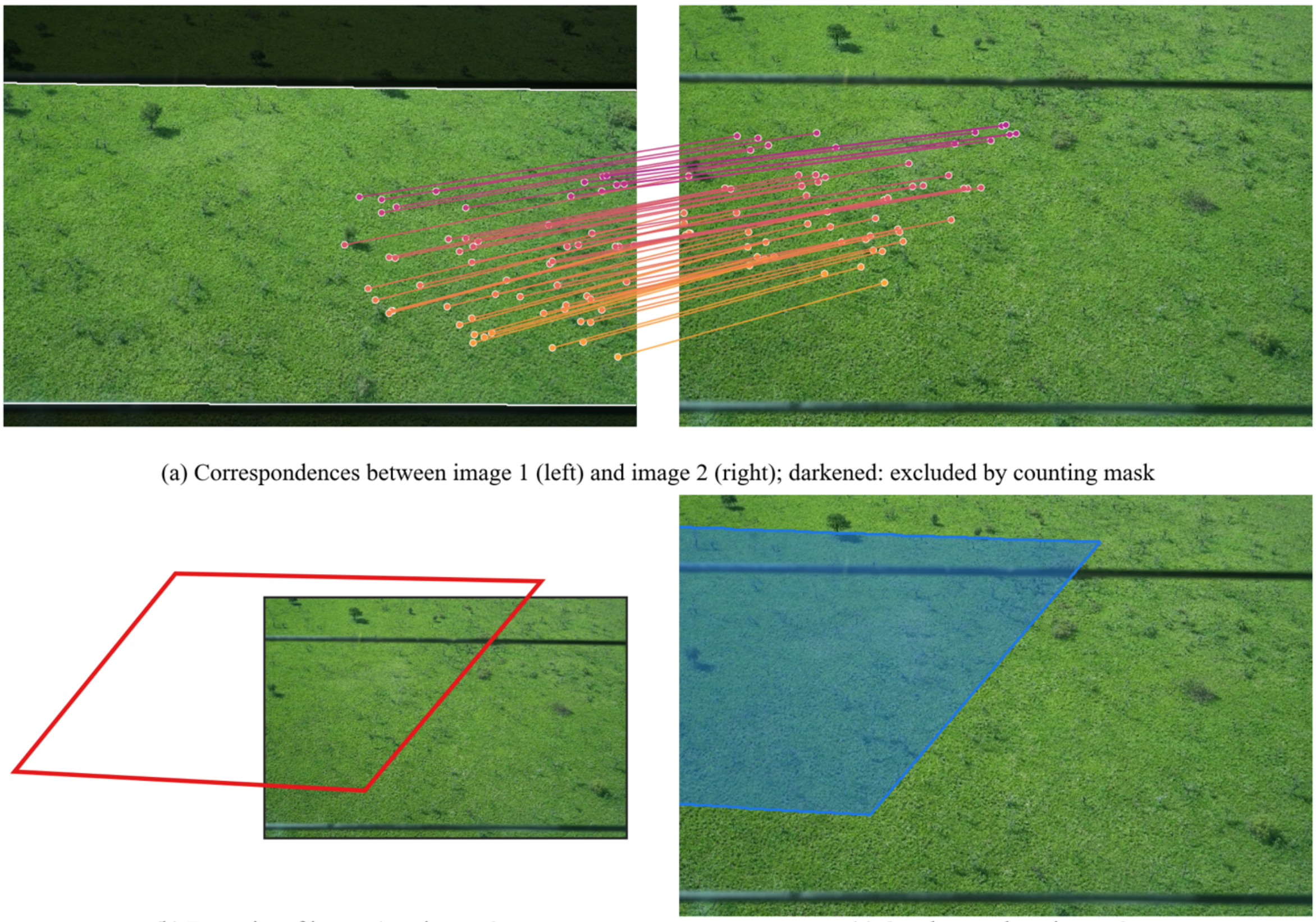

(a) Correspondences between image 1 (left) and image 2 (right); darkened: excluded by counting mask

(b) Footprint of image 1 on image 2

(c) Overlap mask on image 2

**Figure 5.** Overlap extraction between successive aerial images using WildMatch. (a) Selected correspondences between image 1 and image 2; darkened areas indicate regions excluded by the counting mask. (b) Projected footprint of image 1 in the coordinate system of image 2, outlined in red. (c) Geometric overlap region overlaid on image 2 in blue, before restriction to the counting strips. For counting, the exclusion mask is obtained by intersecting this region with both the projected counting strip of image 1 and the counting strip of image 2.

The results in Table 4 show that WildMatch substantially reduces the computational and memory demands of image matching while enabling faster inference even on a CPU. These efficiency gains support its practical use in resource-constrained environments where GPU acceleration is unavailable.

Table 4. Computational cost and CPU inference efficiency of WildMatch and RoMaV2.

| Metric | RoMaV2 (precise) | RoMaV2 (fast) | WildMatch |
|---|---|---|---|
| Number of parameters (M) | 425.4 | 425.4 | 8.05 |
| Weight file size (MB) | 1045 | 1045 | 31 |
| Computational cost per image pair (GFLOPs) | 14,985 | 2,164 | 383 |
| CPU inference time per image pair (s) | 81.00 | 7.44 | 2.69 |
| Peak CPU memory usage (GB) | 15.8 | 4.3 | 2.6 |

Note: "Precise" and "fast" denote the two evaluated RoMaV2 configurations. WildMatch achieves approximately 30.1× and 2.77× CPU inference speedups over RoMaV2 (precise) and RoMaV2 (fast), respectively.

### 5.3 Localization and counting results

WildDet exceeded YOLO11 on the four reported AP and F1 measures (Table 5). The overall mAP was calculated across all five categories. Class-aware average precision increased from 0.774 to 0.785, and best F1 increased from 0.805 to 0.820. Class-agnostic average precision increased from 0.804 to 0.814, and class-agnostic best F1 increased from 0.818 to 0.830. These results describe the detector comparison; they do not quantify the combined effect of strip masking and overlap removal.

Table 5. Animal detection performance. Best F1 values summarize the reported detection comparison.

| Model | mAP | Best F1 | Class-agnostic AP | Class-agnostic best F1 |
|---|---|---|---|---|
| YOLO11 | 0.774 | 0.805 | 0.804 | 0.818 |
| WildDet | 0.785 | 0.820 | 0.814 | 0.830 |

WildDet achieved higher AP than YOLO11 for all four focal species (Table 6). AP increased from 0.795 to 0.802 for Kob, from 0.769 to 0.777 for Gazelle, from 0.905 to 0.922 for Tiang, and from 0.628 to 0.630 for Reedbuck. Tiang had the highest AP and Reedbuck the lowest in both models.

Table 6. Species-level average precision for YOLO11 and WildDet.

| Model | Kob | Gazelle | Tiang | Reedbuck |
|---|---|---|---|---|
| YOLO11 | 0.795 | 0.769 | 0.905 | 0.628 |
| WildDet | 0.802 | 0.777 | 0.922 | 0.630 |

WildDet reduced overall MAE from 0.752 to 0.594 and RMSE from 2.052 to 1.957 relative to YOLO11 (Table 7), corresponding to relative reductions of 21.0% and 4.6%. Mean bias decreased from +0.604 to +0.436, a 27.8% reduction in magnitude. Species-level RMSE decreased for all four focal species: from 1.882 to 1.859 for Kob, 0.300 to 0.284 for Gazelle, 0.559 to 0.508 for Tiang, and 0.373 to 0.276 for Reedbuck. Reedbuck showed the largest relative RMSE reduction (26.0%), whereas Kob retained the highest RMSE in both models.

Table 7. Overall counting errors and species-level RMSE for YOLO11 and WildDet. Counts are obtained from detection boxes; positive mean bias indicates overcounting.

| Model | MAE | RMSE | Mean bias | Kob RMSE | Gazelle RMSE | Tiang RMSE | Reedbuck RMSE |
|---|---|---|---|---|---|---|---|
| YOLO11 | 0.752 | 2.052 | +0.604 | 1.882 | 0.300 | 0.559 | 0.373 |
| WildDet | 0.594 | 1.957 | +0.436 | 1.859 | 0.284 | 0.508 | 0.276 |

## 6. Discussion

This study presents a modular deep learning framework that connects counting-strip delineation, overlap extraction, and animal detection within a common aerial wildlife survey workflow. WildSeg, WildMatch and WildDet address the spatial requirements of the survey protocol and the visual characteristics of aerial imagery through specialized designs. Their integration supports a consistent process for defining observation areas, identifying repeated coverage, and detecting animals within the selected regions. The framework also demonstrates how task-specific choices in image resolution and model architecture can combine high accuracy with the efficiency required for large-scale surveys.

### 6.1 Efficient processing through task-specific image resolution

An important finding is that WildSeg and WildMatch maintain high accuracy when processing resized images. High-resolution photographs contain the fine detail needed to recognize individual animals, but counting-strip delineation and overlap extraction primarily depend on larger structural and geometric relationships. Rods extend across a substantial proportion of the image, while image alignment draws on spatial correspondences distributed throughout the scene. These characteristics allow both tasks to retain useful information after image resolution is reduced.

For WildSeg, resized inputs preserve the overall shape and continuity of the rods, enabling reliable segmentation and geometric boundary fitting. For WildMatch, reduced-resolution images retain sufficient visual information to establish correspondences and estimate the transformation between adjacent frames. The strong performance of both modules demonstrates that processing at the original image resolution is unnecessary for achieving high accuracy in these tasks.

Resizing also substantially improves processing efficiency by reducing the amount of image data passed through the networks. This is especially valuable for dense prediction tasks, where computational demand increases with spatial resolution. The resulting strip masks and overlap regions can be mapped back to the original image coordinates for use with animal detections. This task-specific treatment of resolution provides a practical way to process large aerial image collections efficiently while retaining the spatial information needed for survey counting.

### 6.2 Effectiveness and practicality of rod detection and counting-strip delineation

Aircraft-mounted rods provide a consistent physical reference for defining the observation strip. Their elongated shape and extensive horizontal span offer useful structural information for automated extraction. WildSeg combines an encoder–decoder architecture with learnable offset sampling, allowing feature extraction to accommodate variations in rod position and shape. This design is suited to recovering continuous structures against heterogeneous backgrounds, including areas where vegetation and other landscape features share local visual characteristics with parts of the rods.

The improvement over U-Net highlights the value of adapting segmentation to the geometry of the target. Structural continuity is particularly important for this application because the detected rod edges must form coherent boundaries across the image. Learnable sampling provides flexibility in extracting elongated features, while geometric refinement converts the segmentation output into a spatially consistent representation. These complementary operations connect pixel-level prediction with the observation boundaries required by the survey protocol.

Connected-component filtering uses the expected horizontal extent of the rods to remove shorter background structures, and polynomial fitting generates continuous boundaries from the retained pixels. The resulting masks directly determine which animal detections fall within the counting strip. This approach converts a physical survey reference into a reproducible spatial constraint, supporting consistent image selection across large datasets. The same combination of learned appearance and explicit geometry also provides a basis for adaptation to other survey systems using visible markers to define observation areas.

### 6.3 Accurate and lightweight overlap extraction through knowledge distillation

Repeated coverage between successive photographs creates a direct risk of counting the same animals more than once. WildMatch addresses this by establishing dense correspondences between adjacent images and projecting the preceding counting strip into the current frame. Estimating overlap from image content accommodates changes in repeated coverage caused by aircraft movement and acquisition conditions, allowing the selected counting region to follow the actual relationship between successive photographs.

Dense matching uses visual information distributed across the scene, making it suitable for savanna imagery with varying vegetation structure and unevenly distributed distinctive features. The coarse-to-fine procedure first establishes broader spatial relationships and then progressively refines correspondence locations. Confidence-guided match selection and robust geometric estimation translate these correspondences into a transformation that can be used directly for overlap extraction.

A central advantage of WildMatch is that the student achieves accuracy very close to the RoMaV2 teacher while being substantially lighter and faster. Knowledge distillation transfers correspondence and confidence information from the teacher into a compact architecture, preserving strong task performance at a much lower processing cost. This balance is especially valuable for large-area surveys, where extensive image sequences require matching to be performed repeatedly. Reducing the cost of each image pair makes automated overlap correction more practical at the scale of operational wildlife monitoring.

Restricting correspondence selection to the counting strips further connects matching to the sampling protocol. The intersection of the projected preceding strip and the current strip explicitly identifies repeated coverage within the observation area. Together with WildSeg, this establishes a consistent spatial rule for selecting the regions that contribute to animal counts.

**6.4 Suitability of WildDet for aerial wildlife detection**

Aerial wildlife detection requires sensitivity to small targets while retaining enough context to distinguish animals from vegetation and other background features. WildDet addresses these requirements through dual-scale detection, weighted feature fusion and a dedicated classification pathway. Its improvements in both class-aware and class-agnostic evaluation show that the design benefits detection when species distinctions are retained and when animals are treated as a single category.

Detection heads operating on P3 and P4 concentrate prediction on feature resolutions relevant to small and medium-sized animals. P5 remains available for deeper contextual representation and animal-presence gating. This arrangement focuses detection outputs on the target scales characteristic of the imagery while preserving contextual information within the network. It illustrates the value of adapting a general-purpose detector to the visual properties of a specific survey task.

Weighted feature fusion allows the network to adjust the contributions of features from different levels. This flexibility is relevant when animal visibility changes with apparent size, vegetation cover and background texture. The local classification pathway provides access to backbone features at the detection scales, while bounding-box regression uses fused features. This division accommodates the different information requirements of species recognition and localization within a shared architecture.

The density branch adds supervision related to animal numbers and spatial distribution, complementing the emphasis of bounding-box supervision on individual instances. The empty-image gate introduces another survey-oriented mechanism by using animal presence to determine whether downstream prediction is needed. Together, these components reflect the variation encountered in aerial surveys, where extensive empty areas may alternate with concentrated groups of animals.

**6.5 Interpretation of species-specific detection and counting performance**

The species-level results show that detection and counting improvements take different forms across the four antelope species. Tiang achieved the strongest detection performance and the largest AP gain, whereas Reedbuck showed the greatest relative improvement in counting RMSE. These patterns highlight the value of evaluating individual detection and numerical agreement together, because improvements in one are not necessarily expressed to the same degree in the other.

AP summarizes the precision–recall relationship across confidence thresholds, incorporating both localization and species assignment. Counting error measures the discrepancy between predicted and reference numbers at the evaluated operating setting. These metrics therefore provide complementary perspectives on false positives and missed animals. Reedbuck offers a particularly informative example: its counting RMSE decreased by approximately 26% despite a relatively small AP improvement. For wildlife surveys, this represents a meaningful improvement in numerical agreement.

At the aggregate level, the larger relative reduction in MAE than in RMSE indicates that improvements were more pronounced in average absolute count discrepancy than in the measure that gives greater weight to large errors. The accompanying decrease in positive mean bias shows that predicted counts also moved closer to the reference counts. Together, these changes provide a clearer account of counting performance than detection metrics alone.

Species-level count errors also reflect the number and distribution of animals in the evaluated imagery. Interpreting them alongside AP helps distinguish improvements in individual detection from improvements in predicted numbers. Within the proposed framework, these complementary perspectives connect model performance to the survey objective of obtaining consistent animal counts from explicitly defined observation areas.

## 7. Conclusion

This study presents a unified deep learning framework for wildlife detection and counting in aerial imagery, integrating counting-strip delineation, overlap extraction and animal detection through three complementary modules. WildSeg combines rod segmentation with geometric refinement to define observation strips. WildMatch uses distilled dense image matching to identify repeated coverage without relying on image geolocation. WildDet combines dual-scale detection, weighted feature fusion, density supervision and empty-image gating to support animal detection and counting. Together, these modules connect image analysis with the spatial requirements of aerial wildlife surveys.

Evaluation using aerial imagery of four antelope species in South Sudan demonstrated strong performance across the three tasks. WildSeg improved rod detection accuracy by 6.3% compared with U-Net. WildMatch achieved overlap extraction accuracy only 0.37% lower than that of its teacher while providing a substantially lighter and faster model. Compared with YOLO11, WildDet improved overall AP and best class-aware F1-score by 1.4% and 1.9%, respectively, and reduced detection-based count MAE and RMSE by 21.0% and 4.6%, respectively. These improvements demonstrate gains in both animal detection and counting accuracy.

The framework combines these accuracy gains with design choices suited to large-scale processing. WildSeg and WildMatch maintain high accuracy on resized images, substantially reducing computational demand, while knowledge distillation further improves the efficiency of overlap extraction. The modular organization provides a practical basis for adapting individual components to different survey configurations. By combining explicit observation boundaries, automated overlap correction and improved animal detection, the framework supports consistent and efficient processing of aerial wildlife imagery for conservation monitoring and population surveys.

**Acknowledgments**

The authors thank African Parks Network for providing the Antelope dataset.